%% file: ms.tex
\documentclass{article}

    \usepackage[final]{style/neurips_2020}

\usepackage[utf8]{inputenc} 
\usepackage[T1]{fontenc}    
\usepackage{hyperref}       
\usepackage{url}            
\usepackage{booktabs}       
\usepackage{amsfonts}       
\usepackage{nicefrac}       
\usepackage{microtype}      
\usepackage{xcolor}         
\usepackage[center]{caption}
\usepackage{subcaption}
\usepackage{graphicx}
\usepackage{placeins}

\usepackage[backend=biber,style=numeric,sorting=none]{biblatex}
\graphicspath{ {./images/} }

\title{Reinforcement Learning for Wildfire Mitigation in Simulated Disaster Environments}

\input{sections/authors}

\begin{document}

\maketitle

\begin{abstract}
    \input{sections/abstract.tex}
\end{abstract}

\section{Introduction}
\input{sections/introduction.tex}

\section{Background and Preliminaries}
\input{sections/background.tex}

\section{System Components}
\input{sections/components}

\section{Preliminary Experiments}
\input{sections/experiments}

\section{Discussion}
\input{sections/discussion}

\clearpage
{
\small
\printbibliography
}

\clearpage

\appendix
\section{SimFire}
\label{appendix:simfire}
\input{sections/appendix/simfire}

\section{SimHarness}
\label{appendix:simharness}
\input{sections/appendix/simharness}

\end{document}

%% file: sections/authors.tex
\author{%
Alexander Tapley\thanks{Corresponding author.} \quad Marissa Dotter \quad Michael Doyle \quad Aidan Fennelly \\ \quad \textbf{Dhanuj Gandikota} \quad \textbf{Savanna Smith} \quad \textbf{Michael Threet} \quad \textbf{Tim Welsh}\\
The MITRE Corporation - \textit{AI Security and Perception} \\
\texttt{\{atapley, mdotter, mdoyle, afennelly, dgandikota, sosmith, mthreet, twelsh\}@mitre.org}
}

%% file: sections/abstract.tex
  Climate change has resulted in a year over year increase in adverse weather and weather conditions which contribute to increasingly severe fire seasons. Without effective mitigation, these fires pose a threat to life, property, ecology, cultural heritage, and critical infrastructure. To better prepare for and react to the increasing threat of wildfires, more accurate fire modelers and mitigation responses are necessary. In this paper, we introduce SimFire, a versatile wildland fire projection simulator designed to generate realistic wildfire scenarios, and SimHarness, a modular agent-based machine learning wrapper capable of automatically generating land management strategies within SimFire to reduce the overall damage to the area. Together, this publicly available system allows researchers and practitioners the ability to emulate and assess the effectiveness of firefighter interventions and formulate strategic plans that prioritize value preservation and resource allocation optimization. The repositories are available for download at \url{https://github.com/mitrefireline}.

%% file: sections/introduction.tex
The global effects of climate change such as drought and increased temperatures are exacerbating the frequency and intensity of wildfires \cite{national-climate-assessment} and in-turn increasing the effects of climate change through excessive carbon-dioxide emissions \cite{wired} and significant terrain change. Severe wildfires pose a significant threat to life, property, ecology, cultural heritage, and critical infrastructure - in 2022 alone, over 7.6 million acres of land was burned due to wildfires across the United States \cite{wildfire_acres} costing over 3.5 billion dollars in suppression costs \cite{wildfire_costs}. While wildfires are an essential natural occurrence for maintaining healthy ecological systems \cite{Burchill_2021}, uncontrolled fires, particularly those close to the Wildland Urban Interface (WUI), can present significant risks to public health, life and property, necessitating effective management or suppression measures.

In this paper, we introduce SimFire and SimHarness, a Python-based system to accurately model wildland fire spread and generate appropriate mitigation strategy responses via Reinforcement Learning (RL). SimFire utilizes the Rothermel fire spread formula \cite{rothermel} to simulate the movement of wildland fire through an environment generated from real-world operational or procedurally generated fictional data. Simulated agents can be added to the generated environment and place firefighter-induced mitigations to visualize how wildfires will react to specific mitigation strategies. SimHarness is a machine learning harness designed to train RL agents within SimFire to identify the optimal mitigation strategies for a given wildfire scenario. The combination of SimFire and SimHarness provides a customizable system designed to simulate the spread of wildland fire through a generated environment and suggest optimal mitigation strategies for the given scenario. The repositories are available for download at \url{https://github.com/mitrefireline}.

\subsection{Related Works}

Existing fire spread models \cite{elmfire, quic} and visualization tools \cite{burn3d, pyregence} have brought value to the decision making and planning process for fire chiefs, burn bosses, and land managers. SimFire and SimHarness aims to derive even more insights for planners and mitigators by leveraging agent-based machine learning that identifies the optimal strategies for addressing wildland fire scenarios. In recent years, there has been an increase of academics studying RL for disaster relief and response. In \cite{Hammond, cellular}, both provide open-source RL environments and models for training agents to mitigate the spread of wildfire to limit overall damage. Altamimi \cite{Altamimi} similarly trains an RL agent to mitigate the spread of wildfire through an environment, but does not include open-source code. In all these cases, the environments do not support using real-world data during terrain generation or true fire-spread models such as Rothermel \cite{rothermel}. SimFire aims to fill this gap with realistic environments, research-backed fire spread capabilities, and a design that supports further improvement by the open-source community. Similarly, SimHarness' modular structure makes it compatible with any disaster modeler that utilizes the required \texttt{Simulation} API - not just wildland fire - making SimHarness more flexible than current frameworks and extensible for a variety of disaster scenarios.

%% file: sections/background.tex
\subsection{Rothermel Fire Projection}
The Rothermel surface fire spread model has been widely used in the field of fire and fuels management since 1972 \cite{rothermel}. To model the spread of fire across a given surface, the Rothermel equation takes fuel moisture and wind into account for weather conditions, and slope and elevation into account for terrain conditions. These environmental conditions, weather and terrain, pair with input fuel complexity values to determine the spread of a fire throughout an environment. While Rothermel is considered a valuable tool for estimating the rate of fire spread under certain conditions, its accuracy and applicability can vary depending on factors such as the specific environmental conditions, fuel types, and terrain. Researchers and practitioners often use Rothermel as part of a suite of tools with weather data to better understand and manage wildfires. 

\subsection{Reinforcement Learning}

Reinforcement learning is an agent-based sub-field of machine learning that utilizes a user-designed reward function to train an intelligent agent how to interact within an environment and achieve a desired goal \cite{Sutton1998}. In RL, the environment is defined as a Markov Decision Process \cite{mdp}, MDP, $M = (S, A, P, \rho_0, R, \gamma, T)$, where $S$ is the state space, $A$ is the action space, $P : S \times A \times S \rightarrow [0,1]$ is the state transition probability, $\rho_0 : S \times A \rightarrow [0,1]$ is the initial state probability, $R : S \times A$ is the reward function, $\gamma$ is the discount factor, and $T$ is the maximum episode length. The policy $\pi_\theta\ : S \times A$ assigns a probability value to an action given a state.

Throughout training, the agent receives an observed state from the environment $s_t \in S$, representing the state space information currently available to the agent, and performs an action $a_t \in A$ according to its policy $\pi_\theta$, or, at times, a random policy to encourage exploration. After the agent interacts with the environment via the given action, the environment returns both a next state $s_t^{\prime} \in S$ and reward $r_t \in R$. The agent is trained to find a policy that optimizes the user-defined reward function $R$.

%% file: sections/components.tex
As wildfire frequency and severity increases, it is clear innovation is needed to generate more effective wildfire management and mitigation strategies. SimFire and SimHarness work as a single system to both accurately simulate the spread of a fire through a user-specified environment and suggest mitigation strategies for the given scenario to reduce the fire severity risk and lessen the financial, ecological, and public health impacts that agencies manage.

\subsection{SimFire}
SimFire is an open-source Python tool that simulates realistic wildfire spread over generated environments. These generated environments can be created using procedurally generated fictional data or using topographic and fuel data sources available through LANDFIRE \cite{landfire} to model real-world environments. When using real-world data, users can specify a year to gather data from and an area's GPS coordinates to create a realistic training environment.

SimFire introduces the \texttt{Simulation} class, which is a base class to support a simulated disaster environment. This parent class provides the API necessary for SimHarness to train agents within the environment. \texttt{FireSimulation}, a child class of \texttt{Simulation}, provides a simulated wildfire disaster environment based off the Rothermel equations provided by Andrews \cite{rotherml-eqs} as the basis for the fire spread model. Through a configuration file, users can adjust the simulated environment's size, terrain, fuel sources, and wind dynamics - including complex wind dynamics based off of Navier-Stokes equations \cite{fluid}. SimFire provides a variety of fuel configurations out-of-the-box, including the standard 13 Anderson Behavior Fuel Models \cite{landfire, 13_anderson}, and supports the addition of user-specified fuel types as well. Additionally, users can configure aspects of the fire itself, such as ignition location, rate of spread attenuation, and max fire duration for a single space. The library allows researchers and wildland fire managers control over the scenarios used in their mitigation experiments.

In addition to the fire spread model, SimFire supports the placement of different mitigations to control the spread of fire. Mitigations such as firelines, scratchlines, and wetlines can be placed at any pixel within the simulated environment, allowing users to experiment with different mitigation strategies to see how the fire reacts in certain scenarios. SimFire employs PyGame \cite{pygame}, a scalable and highly-optimized game simulation Python library to visualize the fire spread, agent movements, and agent interactions within the environment. The implemented algorithms and formulas along with the flexibility provided by SimFire allow researchers to define different test scenarios and locations for their mitigation experiments. Additional information about SimFire's fire spread verification, data layers, and agent actions can be found in Appendix \ref{appendix:simfire} along with example fire scenarios.

\subsection{SimHarness}
\label{section:simharness}
SimHarness is a Python repository designed to support the training of RLlib \cite{rllib} RL algorithms within simulated disaster environments. SimHarness takes as input an instance of the \texttt{Simulation} class, such as SimFire's \texttt{FireSimulation}, as the training environment. The \texttt{Simulation} object provides an API that allows SimHarness to move agents around the simulated environment and interact with it by placing mitigations. The \texttt{FireSimulation} agents represent firefighters moving through an environment as a wildfire spreads, placing mitigations such as firelines to limit the spread of the fire within the area.

SimHarness utilizes Hydra \cite{hydra} as a hierarchical configuration management tool to allow users to configure the training parameters of SimHarness. The configuration files provided by SimHarness mirror the structure of the Algorithm Configs used by RLlib for model training, such as \texttt{training}, \texttt{evaluation}, and \texttt{resources}. Users can also configure the parameters used for initializing the Simulation and the agents within the environment. For example, users can configure the \texttt{agent\_speed}, which determines the number of actions an agent can take before the simulation is run, \texttt{interactions}, which are the mitigation techniques an agent can apply to the landscape, and \texttt{attributes}, which determine which attributes are passed as an input dimension to the RL model during training and inference.

Another configurable aspect of the SimHarness environment is the reward function. Users can create a custom reward function for training that emphasizes user-specific goals. This allows for tuning of the agent policy to better suit the user's goals. For example, some users may want policies that prioritize ending the fire as quickly as possible, while others may focus more on limiting the fire spread to specific areas. An example workflow of SimHarness can be found in Appendix \ref{simharness:workflow}.

%% file: sections/experiments.tex
SimFire and SimHarness provide a novel system for generating mitigation strategies for scenarios including real-world data. For this reason, comparisons between SimHarness and other available methods are not one-to-one, but we hope the open-sourcing of SimFire and our preliminary experiments can expand current benchmarks to include the testing of strategies in real-world scenarios.

The following experiment applies the SimFire and SimHarness tools to an area of land in Coalinga, CA near the location of the Mineral Fire of 2020. The fuel and terrain data is the true observed data from 2019 pulled from LANDFIRE \cite{landfire} to simulate the fuels and terrain prior to fire ignition. The area of land simulated covers a 128 unit $\times$ 128 unit square area with a left-corner GPS lat-long location of \texttt{(36.09493, -120.52193)}, where each unit in the grid represents 30 square meters.

In the scenario, the fire starts at a random location within the simulated area and the simulated "firefighter" controlled by a trained DQN \cite{dqn} policy always begins at location ${[0, 64]}$, halfway down the left-hand side of the simulated environment.

\begin{figure}[ht]
    \centering
    \includegraphics[width=.7\textwidth]{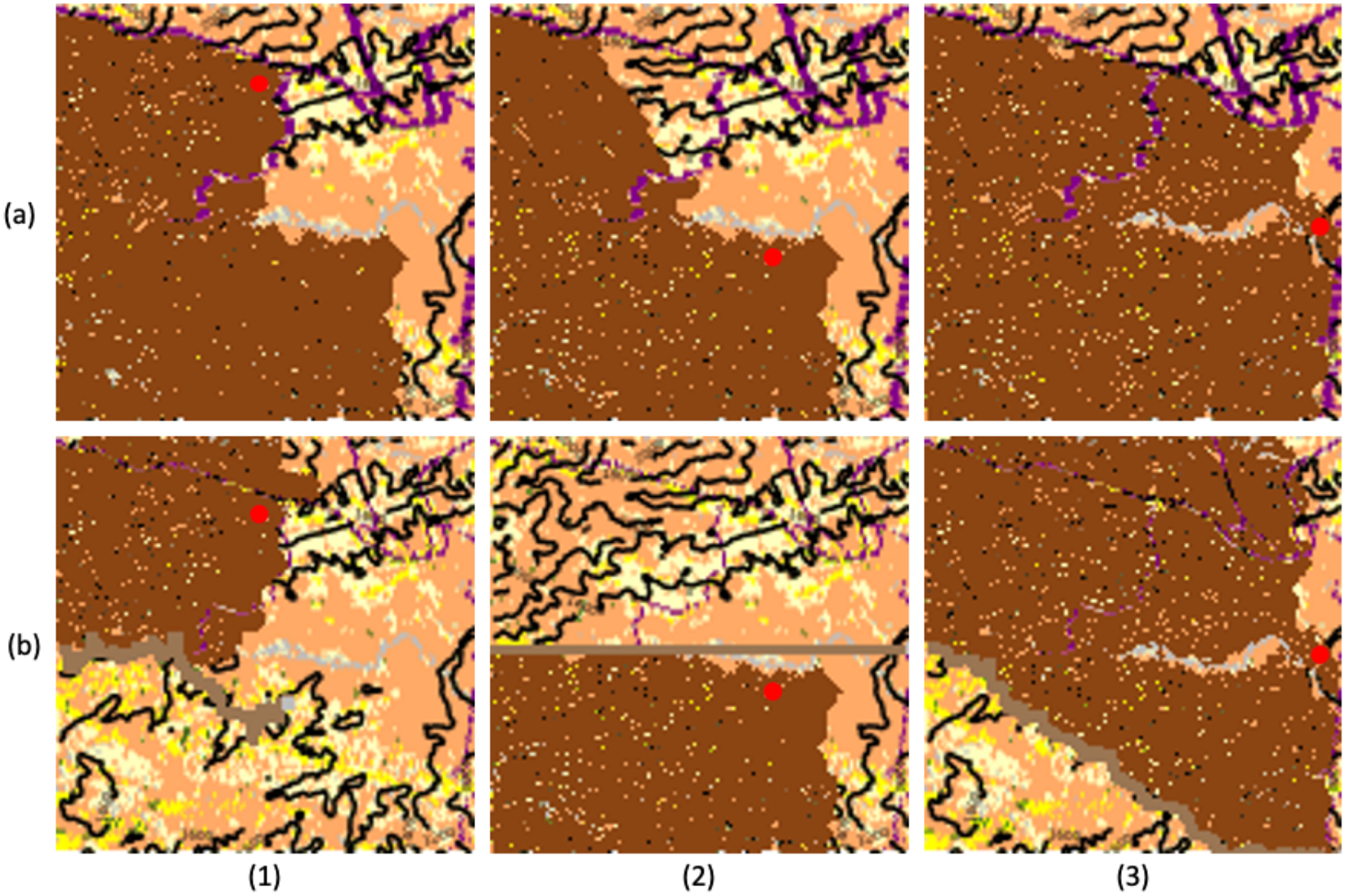}
    \caption{The difference in fire spread for 3 separate scenarios (1, 2, 3) when (a) no mitigations are placed and (b) mitigations are placed using SimHarness. Ignition locations are marked by a red circle.}
    \label{fig:compare}
\end{figure}
\FloatBarrier

As shown in Figure \ref{fig:compare}, the trained agent has generalized to generate successful mitigation strategies for random fire ignition scenarios. In all three cases, the agent's mitigation strategy successfully saved a large section of land from being burned and limited the rate of spread of the fire. In scenario 2, the fire ignites close to latitude of the agent, resulting in the agent needing to move in a straight line to cut the fire off before it can spread to the top half of the environment. The training parameters and quantitative metrics from the training run can be found in Appendix \ref{appendix:simharness}.

%% file: sections/discussion.tex
The SimFire and SimHarness repositories together create an open-source Python-based system to accurately model wildland fire spread and generate appropriate mitigation strategy responses via RL. Researchers and wildfire managers can leverage SimFire and SimHarness to identify new land management strategies that limit the fire severity risk and lessen financial, ecological, and public health impacts caused by increasingly severe wildfires.

In the future, we aim to incorporate additional agent constraints, like agent distance to the fire and realistic movements \cite{nwcg_prod_tables}, to the training process to produce more accurate strategies. We also aim to add agent types and capabilities to more accurately model the range of equipment and crews available to land managers and add data layers to the Simulation to more accurately model the landscape, including buildings and areas of cultural importance such that the environment more accurately models the real-world.

\acksection The authors acknowledge the help of Chris Kempis in the development of SimFire. This work was funded under MITRE's 2022 and 2023 Independent Research and Development Program.

%% file: sections/appendix/simfire.tex
\subsection{Available Data Layers and Actions}
\label{simfire:layers}

SimFire provides Fuel, Terrain, and Wind data layers that are ingested into SimHarness to create the following state spaces at each pixel:

\begin{itemize}
    \item \textbf{Fuel}:
        \subitem \textit{w\_0}: Oven-dry Fuel Load ($lb/ft^2$)
        \subitem \textit{sigma}: Surface-Area-to-Volume Ratio ($ft^2/ft^3$)
        \subitem \textit{delta}: Fuel Bed Depth (ft)
        \subitem \textit{M\_x}: Dead Fuel Moisture of Extinction
    \item \textbf{Terrain}:
        \subitem \textit{elevation}: Elevation (ft) in relation to sea level.
    \item \textbf{Wind}:
        \subitem \textit{wind\_speed}: Wind Speed (mph)
        \subitem \textit{wind\_direction}: Direction of the Wind (degree)
\end{itemize}

These layers are provided to SimHarness as dimensions of the input observation to the RL model. Users can specify which data layers are passed to the model and what order they are within the observation. SimFire also provides the minimum and maximum bounds for each data layer which helps SimHarness normalize observations dimensions, if desired by the user.

The \textit{Fuel} data layer is set based on the type of fuel present in the given scenario, determined by the fire start location and the size of the simulation specified. SimFire supports the usage of the 13 Anderson Behavior Fuel Models \cite{13_anderson}.

SimFire also supports 3 types of firefighting mitigations:
\begin{itemize}
    \item \textbf{Fireline}: Flammable material has been removed by scraping or digging down to mineral soil.
    \item \textbf{Wetline}: Water/suppressant sprayed in an area.
    \item \textbf{Scratchline}: Preliminary, quick-action fireline where flammable material is removed but not entirely and not completely down to mineral soil.
\end{itemize}

Each mitigation has different properties which effect the speed and movement of the fire when the fire is in contact with the mitigation.

\clearpage
\subsection{Additional Fire Scenarios}
\label{simfire:scenarios}

\begin{figure}[ht]
    \centering
    \includegraphics[width=0.9\textwidth]{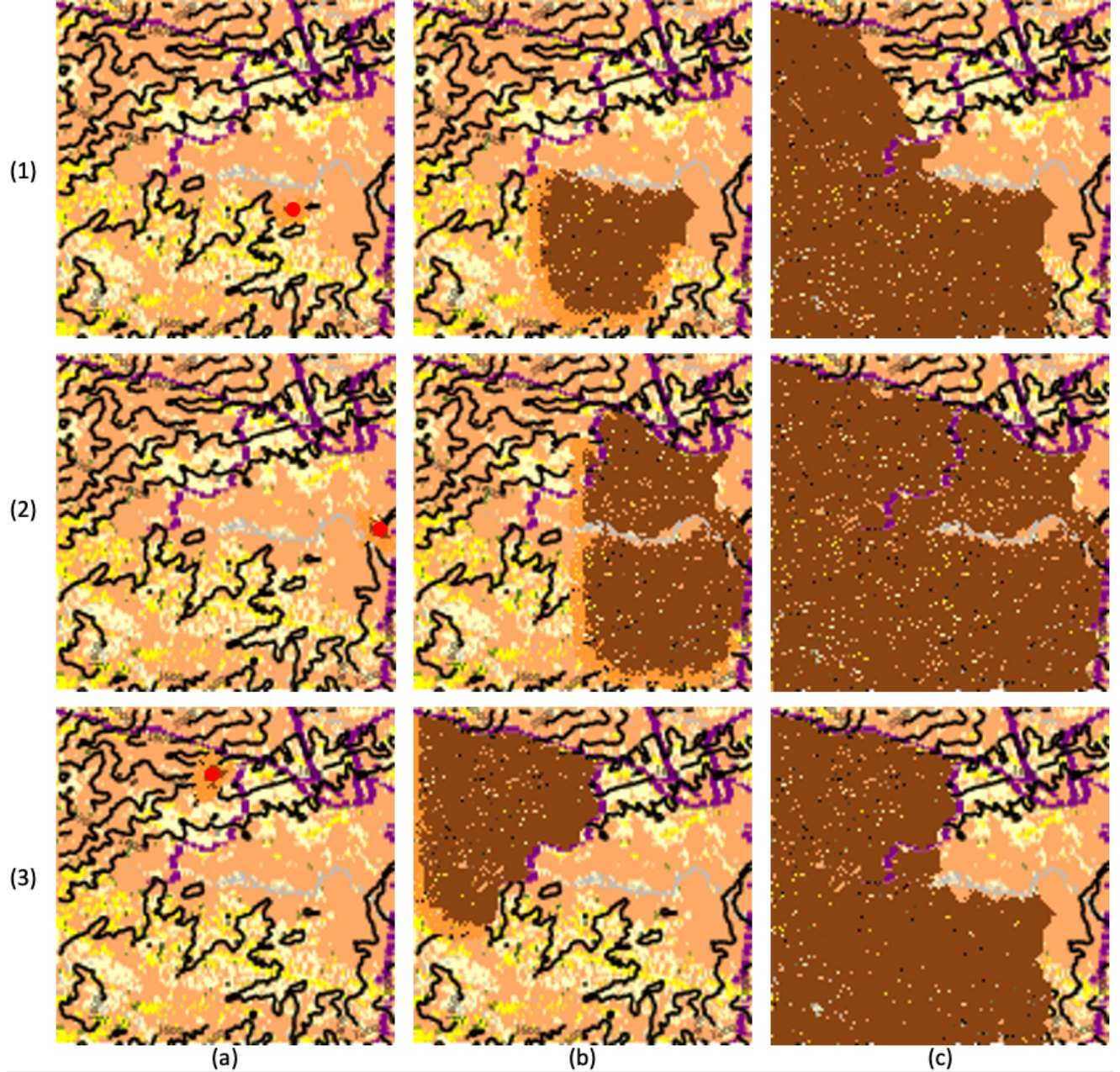}
    \caption{The progression of fire spread in 3 scenarios (1,2,3) using SimFire. Frames are captured at three separate points of the episode: a.) near ignition, b.) mid-episode, and c.) after the fire spread has stopped. Each scenario represents a different fire ignition location, marked by a red circle in frame (a).
}
    \label{fig:fire_scenarios}
\end{figure}
\FloatBarrier

\subsection{Validation and Verification}
\label{simfire:vv}

Future work will provide a detailed comparison of the validation and verification process for the fire spread simulator, SimFire, and the underlying fire spread model, Rothermel to other fire spread models including ElmFire. This study is evaluated using the historical database, BurnMD \cite{dotter}, which includes 308 medium sized fires with near real-time mitigations and daily wildfire perimeters. For more details about the database, its contents, and the data sources, please see the referenced publication or visit the BurnMD website, \url{https://fireline.mitre.org/burnmd}.

%% file: sections/appendix/simharness.tex
\subsection{Workflow}
\label{simharness:workflow}

\begin{figure}[ht]
    \centering
    \includegraphics[width=\textwidth]{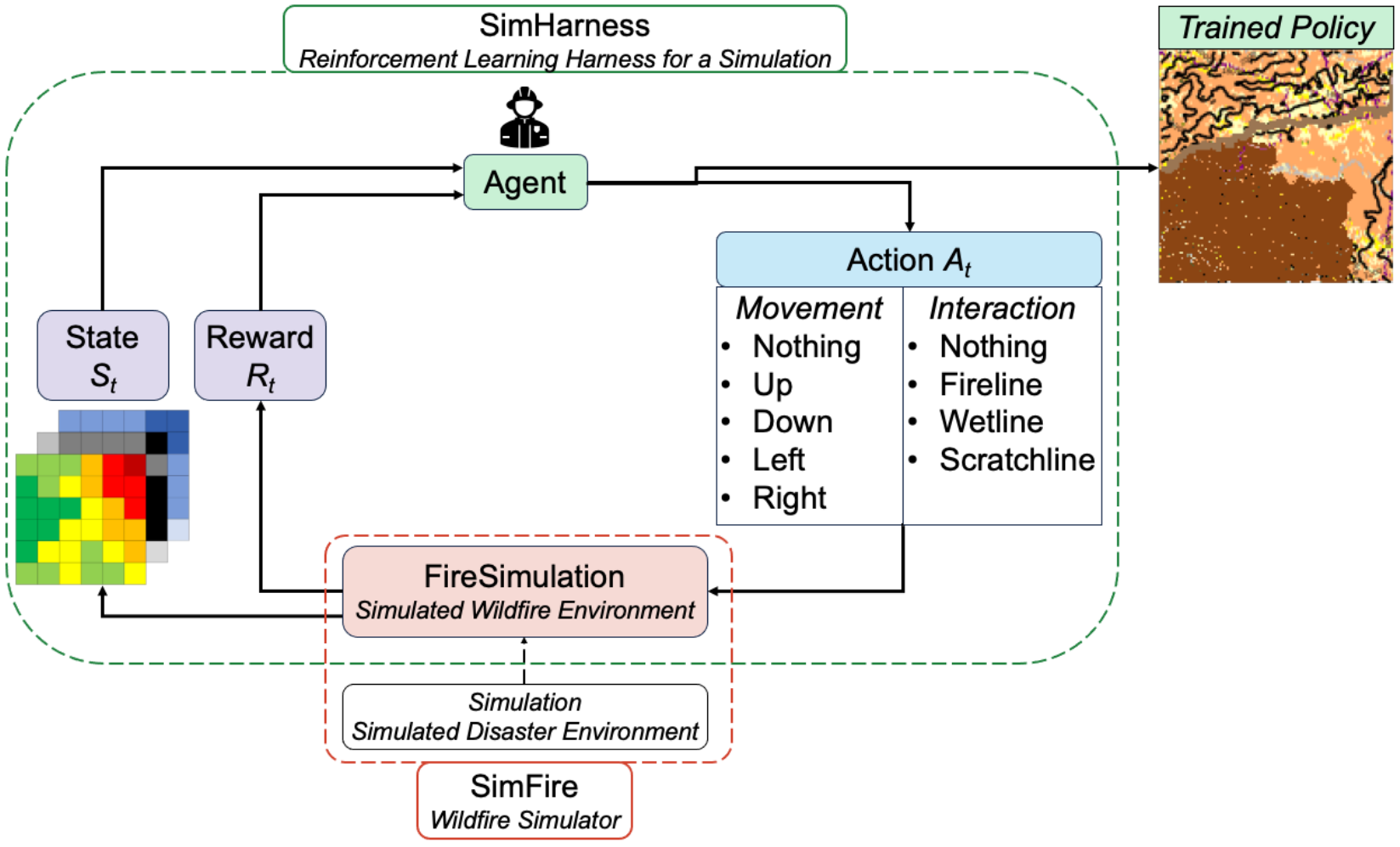}
    \caption{Conceptual workflow for training an RL model using SimHarness within the SimFire environment.}
    \label{fig:workflow}
\end{figure}

SimHarness allows users to train RL agents within any simulated disaster environment, assuming the disaster environment implements the methods required by the \texttt{Simulation} API. In the case of SimFire, SimHarness can generated mitigation strategies for firrefighters to limit the damage caused by wildfires. The general workflow for SimHarness is shown in Figure \ref{fig:workflow}.

\FloatBarrier

The SimHarness training loop functions similarly to a traditional RL training loop, except it expects the passed-in environment to be a child class of \texttt{Simulation} as opposed to a gym \cite{gym} environment. \texttt{Simulation} is currently a class within the SimFire package, but is expected to be moved to a separate, non-disaster-specific package in the future. The simulated environment outputs training signals such as observations and rewards to the SimHarness agent(s) which use the observations to predict optimal actions. The actions produced by the model provide both \textit{movement} and \textit{interaction} information. \textit{Movements} are how the agent is traversing across the environment, such as \texttt{[nothing, up, down, left, right]}. \textit{Interactions} are how the agent is changing the environment itself. In the case of SimFire, this can be \texttt{[nothing, fireline, wetline, scratchline]}. These actions are relayed back to the simulated environment, which then affects the overall disaster scenario simulated by the environment.

\subsection{Training Parameters and Reward Function}
\label{simharness:parameters}

The table below provides a detailed overview of the training parameters leveraged by the environment, agent, learning models, and harness for the experimental results that are presented in Section \ref{section:simharness}. The operational metrics of performance for this experiment are displayed in Section \ref{simharness:metrics}.

A notable aspect of this experiment was the inclusion of a parallel benchmark simulation during each episode. The benchmark simulation simulates the spread of the current wildfire scenario without any agent placed mitigations. The input observation to the model at each step includes 3 arrays: 

\begin{itemize}
    \item Fire Map: Array representing the current fire spread at a given timestep including the placed mitigations and agent locations.
    \item Benchmark Fire Map: Array representing how the fire spread would look at the current timestep if no mitigations were placed.
    \item Final Benchmark Fire Map: Array representing how the fire spread would look at the final timestep if no mitigations were placed.
\end{itemize}

The inclusion of the above information within the training environment is used for both the input observation to the agent as well as the reward function generation, as the reward function compares the total area burned within the mitigated episode to the unmitigated counterpart.

\begin{center}
\begin{tabular}{||c c c||} 
 \hline
 Class & Experiment Parameter & Value \\
 \hline\hline
 Env & Type & Wildfire \\
 \hline
 Env & Simulator & SimFire \\
 \hline
 Env & Type & Operational \\
 \hline
 Env & Left Corner (Lat,Lon) & (36.09493, -120.52193) \\
 \hline
 Env & Geographic Area & 491520 Square Meters \\
 \hline
 Env & Grid Cells & 128x128 \\
 \hline
 Env & Fire Start Location & Random \\
 \hline
 Env & Simulation Data & [Terrain, Fuel, Wind] \\
 \hline
 Agent & Type & Single Agent \\
 \hline
 Agent & Start Location & (64,0) \\
 \hline
 Agent & Observation Space & [current fire map, benchmark fire map, final benchmark fire map] \\
 \hline
 Agent & Action Space Movements & [up, down, left, right] \\
 \hline
 Agent & Action Space Interactions & [Fireline] \\
 \hline
 Agent & Speed & 4 \\
 \hline
 Training & Algorithm & DQN \\
 \hline
 Training & Algorithm Config & [Dueling + Distributional + Noisy] \\
 \hline
 Training & Episodes & 14000 \\
 \hline
 Training & Timesteps & 8273288 \\
 \hline
 Training & Exploration & Epsilon Greedy [1.0, 0.01] \\
 \hline
 Training & Replay Buffer & Episodic Prioritized Replay \\
 \hline
 Training & [lr, gamma] & [0.0045, 0.99] \\
 \hline
 Harness & Type & Ray[Rllib] \\
 \hline
 Harness & CPU & 16 \\
 \hline
 Harness & GPU & 2 \\
 \hline
\end{tabular}
\end{center}

The corresponding reward function for this experiment was based on the incremental proportion of Area saved, or area that was not burned, burning, or mitigated, at each timestep (t), when comparing the mitigated simulation (Sim) to the unmitigated benchmark simulation (Bench). 
\begin{equation}
\label{reward:sim_damaged}
Damaged_{Sim_{t}} =  Sim[Burned_{t}] + Sim[Burning_{t}]+ Sim[Mitigated_{t}]
\end{equation}

\begin{equation}
\label{reward:bench_damaged}
Damaged_{Bench_{t}} = Bench[Burned_{t}] + Bench[Burning_{t}]
\end{equation}

\begin{equation}
\label{reward:sim_new_damaged}
NewDamaged_{Sim_{t}} =  Damaged_{Sim_{t}} - Damaged_{Sim_{t-1}}
\end{equation}

\begin{equation}
\label{reward:bench_new_damaged}
NewDamaged_{Bench_{t}} = Damaged_{Bench_{t}} - Damaged_{Bench_{t-1}}
\end{equation}

\begin{equation}
\label{reward:endangered}
TotalEndangered = Bench_{final_{t}}[Burned]
\end{equation}

\begin{equation}
\label{reward:final}
Reward_{t} = \frac{NewDamaged_{Bench_{t}} - NewDamaged_{Sim_{t}}}{TotalEndangered}
\end{equation}

Equation \ref{reward:sim_damaged} and \ref{reward:bench_damaged} represent the total number of pixels that are "lost" in the simulation at a given timestep, including burned pixels, currently burning pixels, and pixels that have been mitigated by the agent, for the mitigated sim and unmitigated benchmark sim, respectfully. Equation \ref{reward:sim_new_damaged} and \ref{reward:bench_new_damaged} represent the new number of pixels that are "lost" in the simulation at the given timestep. Equation \ref{reward:endangered} is the total number of pixels that will burn if no mitigations are placed in the environment. The final reward in Equation \ref{reward:final} is the difference in new pixels burned between the unmitigated and mitigated simulations as a percentage of the total pixels burned in the benchmark simulation. A positive value represents more pixels being saved in the mitigated scenario than in the unmitigated scenario, with a higher value corresponding to more area saved. A value of 0 means the unmitigated and mitigated scenarios saved the same amount of pixels, and a negative value means the mitigated scenario saved \textit{less} land than the unmitigated scenario. This ensures that the total sum of the rewards within an episode directly corresponds to the total proportion of Area 'saved' for the entire episode (Equation \ref{metric:area_saved}). 

A small positive reward (Equation \ref{reward:addition} is applied to the final reward (Equation \ref{reward:final}) when the agent places a mitigation on an unburned square with no prior mitigations. This addition encourages the agent to place more firelines overall, which helps with training as the agent will get better training examples of how the fire spread reacts to placed mitigations.

\begin{equation}
\label{reward:addition}
Reward_{t} = Reward_{t} + \frac{0.25}{Area_{sim}}
\end{equation}

\subsection{Training Metrics}
\label{simharness:metrics}

The graphs below report the experiment results and the operational metrics of performance for the experiment detailed in Section \ref{simharness:parameters}. The graphs illustrate the Episode Reward Mean, Mean Area Saved, Mean Timesteps Saved, and Mean Burn Rate Reduction over the training of 14,000 episodes.

\begin{figure}[ht]
    \centering
    \includegraphics[width=\textwidth]{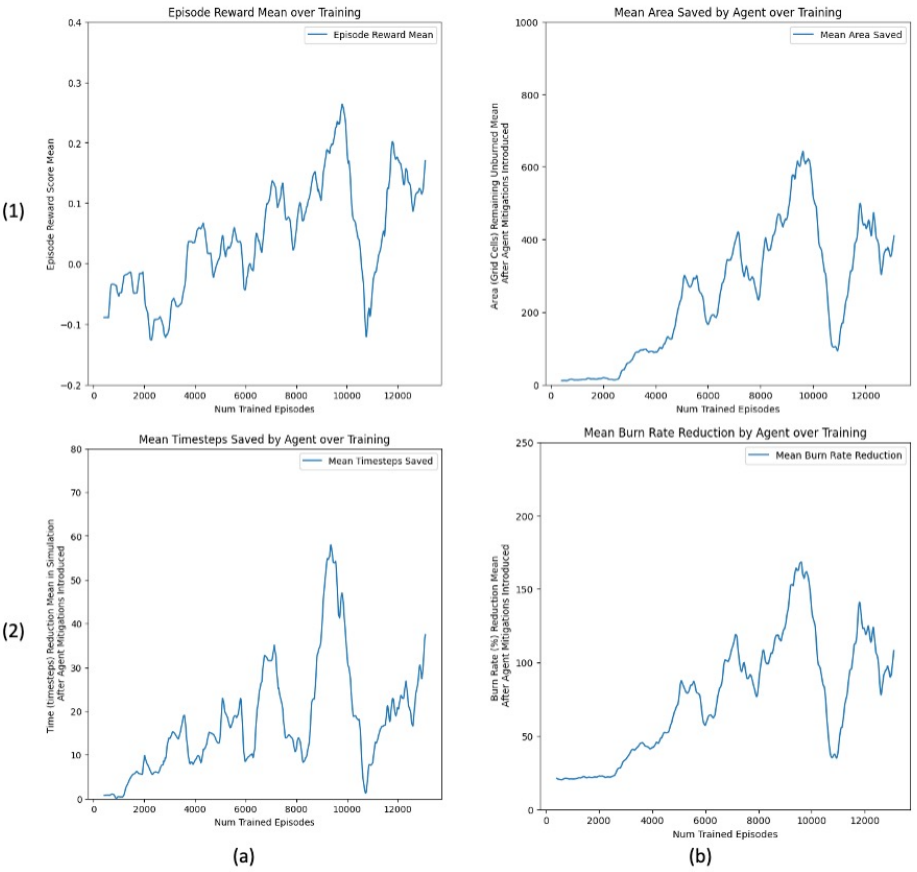}
    \caption{1a: Mean Episode Reward: mean episode received over an episode when mitigations are placed. 1b: Mean Area Saved: mean number of pixels that are left unburned or unmitigated at the end of an episode after mitigations have been placed. 2a: Mean Timesteps Saved: mean number of timesteps that an episode is reduced by when mitigations are introduced compared to the unmitigated scenario. 2b: Mean Burn Rate Reduction: mean reduction in overall episode burn rate (\%) when mitigations are placed compared to the unmitigated scenario.}
    \label{fig:metrics_graphs}
\end{figure}
\FloatBarrier

The metrics per episode (eps) of Mean Area Saved (Equation \ref{metric:area_saved}), Mean Timesteps Saved (Equation \ref{metric:ts_saved}), and Mean Burn Rate Reduction (Equations \ref{metric:burn_rate}, \ref{metric:burn_reduction}) are based on operational metrics that are utilized to estimate wildfire severity and mitigation efficacy. These metrics also serve as heuristic measurements to monitor to validate that the agent is learning effective policies.

\begin{equation}
\label{metric:area_saved}
AreaSaved_{eps} = (Sim[Burned_{eps}] + Sim[Mitigated_{eps}]) - Bench[Burned_{eps}]
\end{equation}

\begin{equation}
\label{metric:ts_saved}
TimestepsSaved_{eps} = Sim[timsteps_{eps}] - Bench[timesteps_{eps}]
\end{equation}

\begin{equation}
\label{metric:burn_rate}
BurnRate_{eps} = \frac{(Burned_{eps} + Mitigated_{eps})}{timesteps_{eps}}*100
\end{equation}

\begin{equation}
\label{metric:burn_reduction}
BurnRateReduction_{eps} = Sim[BurnRate_{eps}] - Bench[BurnRate_{eps}]
\end{equation}